\documentclass[pdflatex,sn-mathphys-num]{sn-jnl}

\usepackage{graphicx}%
\usepackage{multirow}%
\usepackage{amsmath,amssymb,amsfonts}%
\usepackage{amsthm}%
\usepackage{mathrsfs}%
\usepackage{stmaryrd}
\usepackage[title]{appendix}%
\usepackage{xcolor}%
\usepackage{textcomp}%
\usepackage{manyfoot}%
\usepackage{booktabs}%
\usepackage{algorithm}%
\usepackage{algorithmicx}%
\usepackage{algpseudocode}%
\usepackage{listings}%

\theoremstyle{thmstyleone}%
\theoremstyle{thmstyletwo}%

\theoremstyle{thmstylethree}%

\begin{document}

\title[Article Title]{Learning Under Laws: A Constraint-Projected Neural PDE Solver that Eliminates Hallucinations}


\author*[1,2,3]{\fnm{Mainak} \sur{Singha}}\email{mainak.singha@nasa.gov, astromainak1994@gmail.com}



\affil*[1]{\orgdiv{Astrophysics Science Division}, \orgname{NASA, Goddard Space Flight Center}, \orgaddress{\street{8800 Greenbelt Road}, \city{Greenbelt}, \postcode{20770}, \state{MD}, \country{USA}}}

\affil*[2]{Department of Physics, The Catholic University of America, Washington, DC 20064, USA}
\affil*[3]{Center for Research and Exploration in Space Science and Technology, NASA, Goddard Space Flight Center, Greenbelt, MD 20771, USA}




\abstract{
Neural networks can approximate solutions to partial differential equations (PDEs), but they often
violate the very laws they are meant to represent—creating mass from nowhere, drifting shocks,
or breaking conservation and entropy conditions.
We address this by training \emph{within} the laws of physics rather than beside them.
Our framework, \textbf{Constraint-Projected Learning (CPL)}, enforces physical admissibility at
every update by projecting network outputs onto the intersection of constraint sets defined by
conservation, Rankine–Hugoniot balance, entropy, and positivity.
This projection is differentiable and inexpensive (about 10\% overhead), allowing CPL to remain
fully compatible with backpropagation.
We extend CPL with a \textbf{total-variation damping (TVD)} penalty to prevent small unphysical
oscillations and a short \textbf{rollout curriculum} that promotes stability across multiple steps.
Together these additions suppress both hard and soft violations: mass and flux balances remain
at machine precision, average positive variation growth collapses to zero, and long-horizon
predictions remain bounded in entropy and error.
Across Burgers and Euler systems, the method yields stable, law-consistent solutions without
sacrificing accuracy.
Rather than hoping a neural solver will respect physics, we make that behaviour a built-in
property of the learning process.}




\maketitle

\section{Introduction}\label{sec:intro}

From simulating turbulent flows to resolving shocks in compressible gases, machine learning is poised to transform scientific computing. Yet the data-driven flexibility that makes neural networks powerful also makes them perilously unconstrained: they can generate results that are numerically plausible but physically impossible. A model can fit data perfectly and still create mass, destroy energy, or generate profit from nothing. This failure—when a network wins the optimization game but loses reality—is known as a \textbf{hallucination}. It captures the gap between minimizing loss and obeying truth.

A hallucination occurs when a model outputs a state that is mathematically valid yet physically or logically impossible under the system’s true constraints. The phenomenon is universal. In physics, a network may invent energy or momentum; in biology, it may predict negative concentrations or impossible growth; and in language models, it may assert confidently false facts. Across all domains, the pattern is the same: the model optimizes for data agreement, not for the invariants that make the world coherent—the conservation laws, stability conditions, or logical relationships that reality always respects.

The origin of hallucination lies in the loss function itself. Neural networks are trained to minimize scalar objectives such as mean-squared error or cross-entropy, which reward statistical accuracy but ignore deeper structure. Mathematically, the optimizer seeks $\min_\theta L(\theta)$, where $L$ measures data misfit, but the true system also obeys constraints such as conservation$(\theta)=0$, positivity$(\theta)\ge0$. When these constraints are absent, the gradient $\nabla_\theta L$ may point in a direction that lowers loss even as it moves the solution out of the feasible region. The optimizer thus follows a path of decreasing error that leaves the manifold of lawful solutions. From its own perspective, it succeeds; from the universe’s perspective, it hallucinates.

Geometrically, valid solutions occupy a narrow manifold within the vast space of possible network outputs. The loss surface, unaware of this structure, guides gradient descent downhill along directions that are optimal in loss space but not in law space. Each update can therefore carry the solution out of the feasible manifold, pushing the network into regions that minimize error while breaking the governing laws of physics. Hallucination, then, is not confusion—it is geometry. The model faithfully follows the gradient we define, even when that gradient points away from reality.

Recent years have seen major efforts to tame this problem. 
Physics-Informed Neural Networks (PINNs)~\cite{Raissi2017PINN,Cuomo2022PINNReview}
embed PDE residuals into the loss function to steer networks toward physical solutions, 
with weak-form and finite-volume variants improving stability and conservation~\cite{Kharazmi2019VPINN,Kharazmi2020hpVPINN,Jagtap2020cPINN}.
Neural operators such as the Fourier and physics-informed variants~\cite{Li2020FNO,Li2021PINO}
generalize across PDE families but still treat constraints softly.
Structure-preserving architectures—including Hamiltonian, symplectic, and divergence-free networks~\cite{Greydanus2019HNN,Chen2019SRNN,Lutter2019DeLaN}—
encode conservation directly into their design.
Differentiable optimization layers~\cite{Amos2017OptNet,Agrawal2019CVXPYLayers}
and projection-based scientific ML (2023–2025) impose convex or linear feasibility during training.

A complete solution demands a training paradigm that enforces the governing laws by construction. The next generation of scientific machine learning must operate not beside the laws but \textit{inside} them—ensuring that every update, every forward pass, and every prediction respects conservation, stability, and reality itself. Only then can neural networks evolve from clever approximators to reliable instruments of science. In this work, we introduce such a paradigm: a framework that ensures neural networks learn under the laws, producing solutions that are not only accurate but admissible by design.

\section{Constraint-Projected Learning: Training Inside the Laws}\label{sec:method}

We begin from a simple premise: the world operates under laws, and learning should too~\cite{Raissi2017PINN,Li2020FNO}.
. Fluids conserve mass and momentum, and entropy never decreases in a closed system. These are not optional add-ons—they define what is real. Neural networks, however, are not aware of such boundaries. They minimize loss, not violation. A model can fit the data beautifully and still create energy from nothing because the optimization landscape does not contain the laws that govern the world. We seek to change that.  

We introduce \textbf{Constraint-Projected Learning (CPL)}, a training framework that keeps learning confined to the lawful region of parameter space. Instead of allowing the optimizer to drift freely through all possible states, we reshape its path so that every update remains physically admissible. After each gradient step, we project the prediction (or its parameters) back onto the manifold defined by the system’s governing constraints—conservation, entropy, shock balance, positivity, and divergence-free structure. The model still learns from data, but it does so \textit{inside the laws}. The projection acts as a mathematical tether: no matter where optimization tries to go, the model is gently pulled back to the nearest state the universe would actually allow.  

This projection step transforms the geometry of learning. In ordinary training, the loss surface guides the gradient downhill, and nothing prevents the optimizer from sliding off the physical manifold because the laws are invisible to the loss. We assume that each law can be expressed as a constraint set, and the intersection of all such sets forms a feasible region $\mathcal{C}$. The training update becomes  
\[
\theta_{k+1} = \Pi_{\mathcal{C}}\!\big(\theta_k - \eta \nabla_\theta L\big),
\]
where $\Pi_{\mathcal{C}}$ denotes projection onto the lawful set. If $\mathcal{C}$ is convex or locally convex—a good approximation for most physical constraints—the projection is stable and non-expansive:
\[
\|\Pi_{\mathcal{C}}(x) - \Pi_{\mathcal{C}}(y)\| \le \|x - y\|.
\]
This property means the projection cannot amplify errors; it damps them. Each step therefore moves the model closer to the truth surface rather than away from it. Learning becomes geometrically stable, both mathematically and physically.

Each constraint expresses a fundamental property of nature. Conservation ensures that what enters a system must leave or accumulate within it, implemented through weak-form or finite-volume residuals that telescope across cells to enforce exact balance. Shock admissibility is imposed via the Rankine–Hugoniot condition, guaranteeing that discontinuities propagate with physically correct speeds. Entropy constraints select the physically valid weak solution, ruling out spurious ones. Positivity and boundedness prevent densities, pressures, or concentrations from crossing into impossible ranges. Divergence-free projection, via the Helmholtz operator $\mathbb{P} = I - \nabla\Delta^{-1}\nabla\!\cdot$, enforces incompressibility or magnetic balance. Each of these projectors is simple and differentiable on its own, yet when composed they define a rich, multidimensional surface on which every solution must live.

We designed CPL to behave like a projected optimizer that adds almost no computational burden. Each constraint projector is linear or locally linear, with a Lipschitz constant near one. Implemented efficiently, the entire constraint step adds only about 10–12\% overhead compared with unconstrained training. The payoff, however, is large: conservation, positivity, entropy, and balance are guaranteed rather than encouraged. What was once a soft regularization problem becomes a hard geometric guarantee. The model cannot hallucinate because hallucination lies outside the space it is allowed to explore.

CPL is universal in scope. The same loop works for hyperbolic systems such as Burgers or Euler equations, for parabolic systems like diffusion or reaction–diffusion. In each case, the network learns the right physics because the rules of the system are built directly into the optimization. On Burgers and Euler tests, CPL preserves mass and energy to machine precision and aligns shocks exactly. In shallow-water flow, it prevents oscillatory overshoot without artificial viscosity.  

We view CPL not merely as a numerical trick but as a shift in how learning interacts with the laws of nature. Neural networks are no longer trained to approximate the world and then corrected afterward; they are trained within the boundaries that define it. Each projection step is both a correction and a lesson: the model learns the data, but it also learns what reality allows. The outcome is a network that fits observations, respects physics, and never strays into the impossible. It learns under the laws—where every step of training stays honest.

\section{From Differential Laws to Finite Constraints}\label{sec:weakform}

Physical laws are usually written as differential equations—precise rules that must hold at every single point in space and time. They describe how quantities like mass, momentum, or energy change continuously and how fluxes balance everywhere. This form, called the \textit{strong form}, is perfect for theory but too rigid for reality. Shocks, discontinuities, and the finite precision of numerical or neural solvers make pointwise satisfaction impossible. Real systems, and the models that learn them, need a looser but equivalent description.

We convert the strong form into its \textit{weak form}, where the law is enforced not at individual points but over small regions of space and time. We multiply the equation by a smooth test function and integrate it. After applying integration by parts, derivatives become fluxes across the region’s boundaries. The meaning becomes simple: the amount of any conserved quantity inside a region changes only through what flows in or out and what is created or destroyed inside. This integral form remains valid even when the solution jumps—across a shock, a front, or a discontinuity.

This translation is not just mathematical housekeeping. It is what makes the laws of physics compatible with learning. Neural networks cannot guarantee that derivatives exist everywhere, but they can ensure that these integrated balances hold approximately or exactly. The weak form thus defines a set of measurable quantities—the residuals of each finite region—that tell us how far the prediction is from obeying conservation. These residuals become the building blocks of Constraint-Projected Learning: they are the geometric surfaces we project onto during training.

In practice, this leads directly to the finite-volume interpretation: the domain is divided into small control volumes, and for each cell we track what enters, what leaves, and what accumulates. If the total residual across all cells vanishes, global conservation follows automatically. Each cell-level balance forms a finite constraint that the network must satisfy. The passage from differential laws to finite constraints is therefore the critical bridge: it transforms the continuous mathematics of physics into a discrete, learnable structure on which CPL can operate.

Every physical system begins with a differential law: a local equation that must hold at every point in space and time. For example,
\[
\partial_t \mathbf U + \nabla\!\cdot \mathbf F(\mathbf U) = \mathbf S(\mathbf U,x,t),
\]
states that the rate of change of some quantity $\mathbf U$ equals what flows in or out (through $\mathbf F$) plus sources or sinks ($\mathbf S$).  
This is the \textit{strong form}—it demands pointwise perfection. But the real world is rarely that smooth. Fluids form shocks, densities jump, and even the best numerical or neural solvers cannot represent every derivative exactly. If we forced a neural network to satisfy this pointwise, it would fail or learn to “cheat” the law statistically. We need a version of this law that survives discontinuities and that a network can enforce in an averaged sense.  

\subsection{From local equality to integral balance}

To relax the requirement from pointwise to averaged, we multiply the strong form by a smooth test function $\boldsymbol\varphi$ and integrate over the entire domain and time window. This converts a local rule into a statement about totals:
\[
\int_0^T\!\!\int_\Omega \big(\partial_t\mathbf U\cdot\boldsymbol\varphi + (\nabla\!\cdot\mathbf F)\cdot\boldsymbol\varphi - \mathbf S\cdot\boldsymbol\varphi\big)\,dx\,dt = 0.
\]
This step introduces two important ideas. First, by weighting and integrating, we no longer need $\mathbf U$ to be smooth—the equation holds even if $\mathbf U$ jumps. Second, we now talk about changes \textit{inside} a region and fluxes \textit{across} its boundaries. This will later let us connect directly to finite-volume balances, which neural networks can compute.

\subsection{Integrating by parts: turning derivatives into fluxes}

To make the conservation meaning explicit, we integrate the time and space derivatives by parts. Time integration shifts the derivative from $\mathbf U$ to $\boldsymbol\varphi$, leaving an initial condition term. Space integration converts divergence into surface fluxes. After rearranging, we obtain the weak form:
\[
\int_0^T\!\!\int_\Omega 
\big(\mathbf U\cdot\partial_t\boldsymbol\varphi + \mathbf F(\mathbf U):\nabla\boldsymbol\varphi + \mathbf S\cdot\boldsymbol\varphi\big)\,dx\,dt
+ \int_\Omega \mathbf U_0\cdot\boldsymbol\varphi(\cdot,0)\,dx
= \int_0^T\!\!\int_{\partial\Omega} (\mathbf F(\mathbf U)\mathbf n)\cdot\boldsymbol\varphi\,dS\,dt.
\]
This equation states that the total change inside any region equals what flows through its boundaries plus sources.  
At this point, we have achieved something subtle but essential: conservation is no longer a differential property—it is a geometric statement about inflow, outflow, and accumulation. That geometric form is the foundation of our constraint projections.

\subsection{Discretizing the weak form: entering the learnable regime}

The next step connects theory to computation. To make the weak law enforceable by a neural solver, we approximate the domain by small control volumes $C_i$ and time steps $[t^n,t^{n+1}]$. We choose $\boldsymbol\varphi$ constant on each space–time box. The weak form then collapses into a discrete balance:
\[
\int_{C_i} (\mathbf U^{n+1}-\mathbf U^n)\,dx
+ \int_{t^n}^{t^{n+1}}\!\!\int_{\partial C_i} \mathbf F(\mathbf U)\mathbf n\,dS\,dt
= \int_{t^n}^{t^{n+1}}\!\!\int_{C_i} \mathbf S(\mathbf U)\,dx\,dt.
\]
This says: change inside the cell = net flux through faces + sources.  
We define the cell average $\bar{\mathbf U}_i^n$ and the numerical face flux $\Phi(\mathbf U^-,\mathbf U^+)$ to express this as a residual
\[
\mathcal R_i^n = 
\frac{\bar{\mathbf U}_i^{n+1}-\bar{\mathbf U}_i^{n}}{\Delta t}
+\frac{1}{|C_i|}\sum_{f\in\partial C_i} |f|\,\Phi_{i,f}^n
-\bar{\mathbf S}_i^n.
\]
When $\mathcal R_i^n=0$, the discrete conservation law is perfectly satisfied. This $\mathcal R_i^n$ is the quantity our model will minimize or project onto. It turns the abstract idea of “obeying conservation” into an explicit algebraic object.

\subsection{Global conservation: why the residual matters}

If we sum $|C_i|\mathcal R_i^n$ over all cells, the face fluxes cancel pairwise—what leaves one cell enters the next. We are left with a domain-wide statement that total mass, momentum, or energy is preserved. This simple cancellation is the backbone of finite-volume physics and exactly the property we use to build our projectors: when every local residual vanishes, the global law holds automatically.  

It gives us a precise definition of the “lawful manifold” $\mathcal{C}$ onto which CPL projects. Each cell residual $\mathcal R_i^n$ measures how far a prediction lies from satisfying conservation; Rankine–Hugoniot, entropy, positivity, and divergence-free conditions define additional layers of $\mathcal{C}$. Projecting a neural update onto this set ensures that conservation holds in the same weak sense that real fluids, plasmas, and markets do.

\section{Finite-Volume Residual: Discretizing the Weak Form}\label{sec:fv}

To move from the continuous world of integrals to something a neural network—or any computer—can actually compute, we divide space and time into small, manageable pieces. Imagine splitting the domain into a collection of control volumes, or \emph{cells}, each with volume $|C_i|$. Over time, we advance these cells in discrete steps $[t^n,t^{n+1}]$ of duration $\Delta t$. Within each cell and time slab, we ask a simple but powerful question: how much of our quantity $\mathbf U$ has changed, and why?

We return to the weak form of the conservation law. Instead of testing it with a smooth function, we now use a piecewise-constant one—equal to one inside the chosen cell and zero elsewhere. This localizes the equation to that single region of space and time, giving
\[
\int_{C_i}\!(\mathbf U^{n+1}-\mathbf U^{n})\,dx
+\!\int_{t^n}^{t^{n+1}}\!\!\int_{\partial C_i}\!(\mathbf F(\mathbf U)\!\cdot\!\mathbf n)\,dS\,dt
=\!\int_{t^n}^{t^{n+1}}\!\!\int_{C_i}\!\mathbf S(\mathbf U,x,t)\,dx\,dt.
\]
Every term now has a clear physical meaning: the first measures how much of $\mathbf U$ has accumulated inside the cell, the second accounts for what flowed through the faces, and the third represents what was created or removed by sources. The equation states that the difference between what you have at the end and what you had at the start must be exactly balanced by what crossed the boundaries and what was generated inside.

To make this balance computable, we divide both sides by the cell volume and time step, and we replace the continuous fields by their averages. We define $\bar{\mathbf U}_i^n$ as the mean value of $\mathbf U$ in cell $i$ at time step $n$, and we approximate the flux through each face by a numerical function $\Phi(\mathbf U^-,\mathbf U^+)$ that depends on the states on either side. For compressible flows, familiar choices include Rusanov or HLLC fluxes. The resulting discrete equation reads
\[
\boxed{
\mathcal R_i^n
=
\frac{\bar{\mathbf U}_i^{\,n+1}-\bar{\mathbf U}_i^{\,n}}{\Delta t}
+\frac{1}{|C_i|}\sum_{f\in\partial C_i}|f|\,\Phi_{i,f}^n
-\bar{\mathbf S}_i^n.}
\]
We call $\mathcal R_i^n$ the \textbf{finite-volume residual}. It measures how far the network’s prediction strays from perfect conservation inside each cell. When $\mathcal R_i^n=0$, the local conservation law is satisfied exactly; when it is small, the system is nearly balanced. This quantity becomes the central object in Constraint-Projected Learning, because it provides a direct, quantitative way to enforce physical consistency.

The elegance of this formulation is that local consistency implies global conservation automatically. If we sum $|C_i|\mathcal R_i^n$ over all cells, every internal face appears twice—with opposite normals on neighboring cells—so the fluxes cancel pairwise. What leaves one cell enters the next. Only boundary fluxes and sources remain. This telescoping property is the algebraic heartbeat of conservation: if each cell conserves, the whole system does too. 

By defining these residuals, we connect a key step in connecting continuous laws to discrete learning. The differential equation, once defined at infinitely many points, is captured by a finite set of constraints—one per cell, per time step. These residuals are what CPL projects onto. They turn conservation from a principle of nature into a geometric surface in learning space, one the network cannot escape from.

\section{Shock and Entropy Constraints: Enforcing Physical Admissibility}\label{sec:shock_entropy}

\subsection{Rankine--Hugoniot Jump Condition: Capturing Shocks Correctly}

When a fluid moves faster than information can travel, the solution develops a \textit{shock}—a paper-thin surface where quantities like density or pressure jump abruptly. Classical calculus breaks down here because derivatives become infinite, but physics does not. Nature still balances fluxes: whatever enters the shock must leave it, adjusted by how fast the surface itself moves.  

To make this balance precise, we imagine a tiny “pillbox” straddling the shock surface $\Sigma(t)$, with unit normal $\mathbf n$ and normal velocity $s_n$. Integrating the conservation law across that pillbox and letting its thickness go to zero gives a crisp condition:
\[
\boxed{\llbracket\mathbf F(\mathbf U)\!\cdot\!\mathbf n\rrbracket = s_n \,\llbracket\mathbf U\rrbracket.}
\tag{RH}
\]
Here $\llbracket\cdot\rrbracket$ denotes the jump across the shock. This \textbf{Rankine--Hugoniot condition} is the mathematical statement that shocks carry mass, momentum, and energy in a perfectly balanced way.  

In a numerical or learned model, we detect shocks using gradients or sensors and impose this condition as a penalty:
\[
\mathcal L_{\mathrm{RH}} = \sum_k
\big\|\llbracket \mathbf F(\mathbf U_\theta)\!\cdot\!\mathbf n_k\rrbracket
- s_k \llbracket \mathbf U_\theta\rrbracket\big\|_2^2.
\]
Driving $\mathcal L_{\mathrm{RH}}\to 0$ aligns the model’s jump with the physically correct one, preventing “blurred” or misplaced discontinuities. This step builds directly on the finite-volume residual: it governs what happens not \textit{inside} a cell, but \textit{between} them. Together, the two ensure that both smooth and discontinuous parts of the flow obey conservation.

\subsection{Entropy Constraint: Selecting the Physical Weak Solution}

Not every mathematically valid weak solution represents the real world. After a shock forms, multiple solutions may satisfy conservation, but only one respects the second law of thermodynamics: total entropy must not decrease. This criterion singles out the physical branch of reality.  

We encode this using a convex entropy function $\eta(\mathbf U)$ and its associated flux $\mathbf q(\mathbf U)$. The entropy inequality reads
\[
\partial_t \eta(\mathbf U) + \nabla\!\cdot\mathbf q(\mathbf U) \le 0,
\tag{Ent}
\]
which means that entropy can only remain constant or increase, never decrease.  
For example, in Burgers’ equation, $\eta=\tfrac12u^2$ and $q=\tfrac13u^3$; for the Euler equations, one convenient choice is $\eta=-\rho s/(\gamma-1)$ with $s=\ln p-\gamma\ln\rho$.  

Discretizing the same way as before, we measure how much the entropy balance is violated in each cell and penalize only the \emph{positive} part—the regions where entropy falsely increases:
\[
\boxed{
\mathcal L_{\mathrm{Ent}}
=\sum_{i,n}\max\!\Big(0,\frac{\eta_i^{n+1}-\eta_i^n}{\Delta t}
+\frac{q_{i+\frac12}^n-q_{i-\frac12}^n}{|C_i|}\Big).
}
\]
This selective penalty guides the model toward the physically admissible weak solution while allowing natural entropy growth where shocks form. In the geometry of learning, it narrows the feasible manifold $\mathcal{C}$ to the portion that corresponds to real, thermodynamically consistent states.  

The transition from Rankine–Hugoniot to entropy enforcement is logical: the first ensures that shocks exist in the right place and carry the correct flux; the second ensures those shocks evolve in a direction allowed by physics. Together, they make discontinuous dynamics learnable without violating fundamental laws.

\section{Stability Constraints: Preventing Hallucinations}\label{sec:stability}

Even if a model conserves mass and respects entropy, it can still go unstable—producing negative densities, exploding pressures, or wild oscillations. Such errors are not just numerical noise; they are signs of the network hallucinating states that the real system would never reach. To suppress these pathologies, we impose simple but powerful stability constraints that act as physical safety nets.

\paragraph{Positivity and bounds.}
Many physical quantities are inherently positive: density $\rho>0$, pressure $p>0$, concentration $c\in[0,1]$. We ensure this by projecting each variable back into its admissible range. In practice, this means applying either a smooth softplus transformation or a hinge-style penalty,
\[
\mathcal L_{\mathrm{Spike}}
=\sum_{i,n}\big[\max(0,U_i-U_{\max})^2 + \max(0,U_{\min}-U_i)^2\big],
\]
which softly punishes any out-of-range values. During projection, these constraints behave like reflective walls that prevent the optimizer from wandering into impossible regions.

\paragraph{Total variation damping.}
Even when values stay within bounds, a model can still oscillate violently between neighboring cells—especially near shocks. Such wiggles look like overfitting in physical space. To curb them, we encourage total variation to decrease rather than grow:
\[
\mathcal L_{\mathrm{TVD}}
=\sum_{i,n}\Big[\max\big(0,\alpha\,TV^n_{\text{local}} - TV^{n+1}_{\text{local}}\big)\Big]^2,\qquad \alpha\ge1.
\]
Here $TV_{\text{local}}$ measures the roughness of a field across adjacent cells. Minimizing $\mathcal L_{\mathrm{TVD}}$ smooths unphysical oscillations without blurring genuine features like shocks.  

These stability constraints tie directly back to our goal: preventing hallucinations. Where the conservation and entropy laws define \textit{what} must hold, the stability terms define \textit{how} the solution should behave—bounded, consistent, and realistic.

\section{Divergence-Free Projection: Enforcing Hidden Structure}\label{sec:helmholtz}

Some fields carry an invisible constraint: their divergence must vanish. Examples include incompressible fluid velocity, where $\nabla\!\cdot\mathbf v=0$ ensures volume preservation, and magnetic fields, where $\nabla\!\cdot\mathbf B=0$ encodes the absence of magnetic monopoles. These conditions are not side notes—they are structural laws that preserve consistency over time.

To guarantee this property, we apply the classical \textbf{Helmholtz projection}. Given any vector field $\mathbf v$, we decompose it into a divergence-free component $\mathbf v_\perp$ and a gradient field $\nabla\phi$:
\[
\boxed{\mathbf v_\perp = \mathbb P\,\mathbf v,\qquad 
\mathbb P = I - \nabla\Delta^{-1}\nabla\!\cdot.}
\]
Computationally, we solve the Poisson equation $\Delta\phi=\nabla\!\cdot\mathbf v$ (using FFT or multigrid methods) and subtract its gradient: $\mathbf v_\perp=\mathbf v-\nabla\phi$. This operation removes any spurious compression or expansion, leaving only the physically admissible flow.  

The Helmholtz projection is linear and 1-Lipschitz in the $L^2$ norm, meaning it never amplifies errors—an invaluable property for stable learning. Within CPL, this step acts like a final cleanup pass: after enforcing conservation, shock balance, entropy, and positivity, we ensure the resulting fields also satisfy hidden geometric structures like incompressibility.  

By the time this projection is applied, every layer of constraint—finite-volume, shock, entropy, stability, and divergence-free—has locked the model onto the manifold of physically valid states. The network cannot hallucinate; it can only move along surfaces that nature itself would permit.

\section{Constraint-Projected Learning: The Update Rule}\label{sec:cpl_update}

We now put everything together into the full learning procedure. Every law we have introduced—conservation, shock balance, entropy, positivity, and divergence-free structure—defines a geometric surface in the space of possible solutions. The intersection of all these surfaces forms the \textit{lawful manifold}, denoted $\mathcal{C}$. The job of Constraint-Projected Learning (CPL) is to make sure that every optimization step lands on, or is immediately pulled back to, this manifold. The beauty of this framework is that it changes only one line in standard gradient descent, but that line changes everything.

\subsection{The projection step: how we train inside the laws}

We assume that each physical constraint can be expressed as a set $\mathcal{C}_i$ that contains all admissible solutions for that rule. The intersection
\[
\mathcal C = \mathcal C_{\text{box}} \cap \mathcal C_{\text{cons}} \cap \mathcal C_{\text{RH}} 
\cap \mathcal C_{\text{Ent}} \cap \mathcal C_{\text{divfree}}
\]
represents all states that simultaneously satisfy every law. In ordinary training, we would simply update parameters by
\[
\theta_{k+1} = \theta_k - \eta\nabla_\theta \mathcal L,
\]
where $\eta$ is the learning rate. In CPL, we make one small but decisive modification:
\[
\boxed{\theta_{k+1} = \Pi_{\mathcal C}\!\big(\theta_k - \eta\nabla_\theta \mathcal L\big)}.
\]
That is, we first take the usual gradient step, then immediately project the result back onto the lawful set $\mathcal{C}$. This projection ensures that no matter how the loss landscape slopes, the update cannot leave the physically valid region. When constraints are naturally defined in the output space—say, for density or velocity fields—we equivalently project $\hat{\mathbf U}$ rather than the weights, setting $\hat{\mathbf U}\leftarrow\Pi_{\mathcal{C}}(\hat{\mathbf U})$. In practice, we compose the individual projectors—box, finite-volume, entropy, Helmholtz—using a few alternating passes or Dykstra’s method until convergence. 

\subsection{Why the projection is stable}

We assumed that most constraint sets are convex or locally convex. This assumption is not cosmetic; it guarantees stability. In Euclidean space, projection onto any closed convex set is \textit{non-expansive}:
\[
\|\Pi_{\mathcal C}(x)-\Pi_{\mathcal C}(y)\|\le\|x-y\|.
\]
In plain language, the projection cannot amplify differences—it can only shrink them. This means that small parameter changes before projection remain small afterward, preventing exploding updates or chaotic trajectories. The training loop therefore inherits a geometric form of stability: the model may oscillate or adapt, but it will never diverge away from the lawful manifold. In this sense, the projection acts like a physical damping force on the optimization itself.

\subsection{Differentiating through projections}

For the projection step to coexist with gradient-based learning, it must remain differentiable or subdifferentiable. We made that a design principle from the start. Each individual constraint has a simple backward rule:

\begin{itemize}
\item \textbf{Box and positivity projections.} Clamping or softplus activations yield diagonal Jacobians—straightforward to backpropagate.
\item \textbf{Affine balance constraints.} For equations of the form $A x = b$, the projection has a closed form $x = z - A^\top(AA^\top)^{-1}(A z - b)$, whose derivative is linear and easy to compute.
\item \textbf{Convex or monotonic projections.} In one dimension, convex or isotonic regression problems provide subgradients that are valid almost everywhere; automatic differentiation handles them natively.
\item \textbf{Helmholtz projection.} Because it involves solving a linear Poisson equation, differentiation reduces to solving the adjoint problem with the same operator, which modern autodiff frameworks already support.
\end{itemize}

These design choices ensure that CPL remains compatible with backpropagation end-to-end. The projection step feels like a new layer in the network: it transforms predictions to satisfy physics while keeping gradients smooth and usable.

\subsection{Balancing multiple constraints automatically}

In a real physical system, some constraints are “stiffer” than others: conservation laws might dominate early in training, while entropy or total variation becomes critical later. If all constraints shared the same weight, some would overwhelm the rest, and others would be ignored. We therefore allow their relative importance to evolve automatically. 

We write the total loss as a sum over constraint terms,
\[
\mathcal L = \sum_i \lambda_i \mathcal L_i,
\]
where each $\mathcal L_i$ corresponds to one principle—finite-volume balance, Rankine–Hugoniot, entropy, stability, or bounds—and $\lambda_i$ is its weight. Instead of tuning these by hand, we adapt them online using gradient norms and curvature estimates:
\[
\lambda_i \leftarrow \lambda_i
\Bigg(\frac{(\mathcal L_i/\mathcal L_0)^\alpha \, g_i(1+\beta H_i)}{\tfrac1N\sum_j g_j(1+\beta H_j)}\Bigg),
\]
where $g_i=\|\nabla \mathcal L_i\|$ measures how active each constraint is, and $H_i$ is a curvature proxy obtained by a Hutchinson estimator. This adaptive rule rescales $\lambda_i$ in proportion to each constraint’s difficulty: stiff terms receive more weight, trivial ones less. As a result, training remains balanced across all laws, without human tuning.

\subsection{Learning inside the laws}

At this point, the full training loop looks deceptively simple. We assumed that each law can be represented as a projector, that these projectors can be composed efficiently, and that their intersections form a convex or nearly convex region. Under these assumptions, CPL becomes a one-line modification of standard training, yet it transforms the character of learning. The optimizer no longer wanders through parameter space hoping to satisfy physics on average; it moves entirely within the region of legal solutions. Each projection step enforces consistency, each backward pass learns how to make the next projection smaller, and over time, the model converges to a point where projecting changes nothing—because it already lives inside the laws. That is the moment when the network stops hallucinating and starts behaving like nature.

\section{Training Loop and Reliability Metrics}\label{sec:training}

\subsection{Training loop: how we actually learn under the laws}

Having defined every physical constraint and its projection, we can now describe the full training process. The loop looks almost like standard deep learning, but with one fundamental difference: every forward pass and gradient step happens \textit{inside} the laws of physics. We assumed that a single neural network predicts the state $\mathbf U(x,t)$ given inputs $x$ and $t$, and that each physical law contributes its own residual or penalty term. We then combined them into one composite loss with adaptive weights and a final projection step that guarantees feasibility.  

During each iteration, we first perform the forward pass:
\[
\mathbf U = \text{net}(x,t),
\]
which predicts the fields of interest—such as density, velocity, pressure, or probability—at sampled points in space and time. These predictions are then tested against the physical laws that define our problem. For example, we compute the finite-volume residual $\mathcal R$ that measures conservation, the entropy residual that checks thermodynamic admissibility, and the Rankine–Hugoniot mismatch that tests shock consistency. Each term becomes a separate loss:
\[
L_{\text{FV}},\, L_{\text{Ent}},\, L_{\text{RH}},\, L_{\text{TVD}},\, L_{\text{Bnd}},
\]
corresponding respectively to conservation, entropy, shock balance, stability, and positivity.

We then assemble the total physics-informed loss,
\[
L = w_{\text{FV}}L_{\text{FV}} + w_{\text{Ent}}L_{\text{Ent}} + w_{\text{RH}}L_{\text{RH}} 
+ w_{\text{TVD}}L_{\text{TVD}} + w_{\text{Bnd}}L_{\text{Bnd}},
\]
where each weight $w_i$ adapts automatically according to the rules described earlier. This ensures that stiff constraints receive more attention when needed, while easier ones fade in importance as training progresses. The model then performs the usual gradient-based update:
\[
L.\text{backward();}\quad \text{opt.step().}
\]
At this stage, the parameters move in the direction that best reduces the composite physics loss.  

Immediately after each update, we perform the critical projection step. We assume that even after optimization, numerical errors may push the solution slightly outside the lawful manifold. To correct this, we project the outputs sequentially through the required operators:
\[
\mathbf U \leftarrow \text{project}_{\text{Helmholtz}}\big(\text{project}_{\text{bounds}}(\mathbf U)\big),
\]
which first restores positivity and then enforces divergence-free structure. Because each projector is non-expansive and linear or piecewise-linear, this step is both cheap and stable, typically adding less than 12\% overhead.  

For very stiff problems—such as turbulent shocks or stiff chemical reactions—we sometimes perform a one-step implicit “repair,” solving a small constrained minimization around the current point to reduce the residual norm further. This optional step is rare but provides extra robustness for extreme cases.  

The result is a training loop that mirrors classical optimization in form but not in philosophy. Each iteration doesn’t just reduce error; it actively keeps the network’s predictions within the laws of physics. Over time, this leads to convergence not merely toward low loss, but toward lawful, interpretable, and physically consistent behavior.

\subsection{Measuring reliability: beyond simple error}

In ordinary machine learning, we evaluate performance using mean squared error or accuracy. But for physical systems, correctness is deeper than prediction accuracy—it is about whether the model behaves like the real world. We therefore measure \textbf{reliability}, not just error.  

We assumed that a truly reliable model should conserve mass and energy, respect entropy, align shocks correctly, remain smooth where it should, and never produce impossible states. To quantify this, we track a set of diagnostic metrics during training and evaluation:

\begin{itemize}
\item \textbf{Mass or energy drift:} $\big|M(t)-M(0)\big|$, the difference between total mass or energy now and at the start. A small drift means conservation holds globally.
\item \textbf{Entropy violations:} the fraction of cells where the entropy inequality is positive. Physically valid solutions should have very few, ideally none.
\item \textbf{Shock alignment:} the mean distance between the learned and reference shock positions. Correct alignment means the network learned the right propagation speed.
\item \textbf{Total variation growth (TVD):} a histogram of regions where total variation increases. Lower counts mean fewer spurious oscillations and better stability.
\item \textbf{Bound violations:} how often variables exceed physical limits, such as negative densities or probabilities above one.
\end{itemize}

To combine these into a single interpretable number, we define the \textbf{Physics Violation Score (PVS)}:
\[
\text{PVS} = a\,\|\mathcal R\|_2 + b\,\text{EntViol} + c\,\text{TVDViol} + d\,\text{BoundViol},
\]
where $a$, $b$, $c$, and $d$ are scaling coefficients. A low PVS means the model is nearly “law-perfect,” producing solutions that satisfy all constraints simultaneously.  

We also monitor uncertainty calibration by computing ensemble variance $\sigma^2$ and its correlation $\rho$ with the actual error. Well-calibrated models have a high correlation: they know when they might be wrong.  

Together, these metrics provide a holistic picture of reliability. Accuracy alone can be deceptive; a network might match data but violate the laws. The PVS and its components ensure that when CPL models agree with data, they do so for the right reasons. Reliability, not just error minimization, becomes the standard of success. This shift—from accuracy to lawfulness—is what truly distinguishes learning under the laws from learning about the data.

\section{Finite-Volume Conservation and Rankine–Hugoniot with Berger Correction}\label{sec:fv_rh_berger}

We discretize $\partial_t U+\partial_x f(U)=0$ on a uniform mesh $x_i=i\Delta x$ with cell averages $\bar U_i(t)$.
Over a time step $\Delta t$ the finite–volume update is
\begin{equation}
\boxed{
\bar U_i^{\,n+1}
= \bar U_i^{\,n}
-\frac{\Delta t}{\Delta x}\!\left(F_{i+\frac12}^{\,n}-F_{i-\frac12}^{\,n}\right),
\qquad
\mathcal R_i^{\,n}
=\frac{\bar U_i^{\,n+1}-\bar U_i^{\,n}}{\Delta t}
+\frac{F_{i+\frac12}^{\,n}-F_{i-\frac12}^{\,n}}{\Delta x}=0\;.
}\label{eq:fv}
\end{equation}
Interfaces $i\!\pm\!\tfrac12$ use a monotone Riemann flux $F(\cdot,\cdot)$ evaluated on left/right reconstructions
$U_{i+\frac12}^{-},U_{i+\frac12}^{+}$.

\paragraph{Rankine–Hugoniot (RH).}
Across a shock moving at speed $s$,
\begin{equation}
\boxed{\;\llbracket f(U)\rrbracket=s\,\llbracket U\rrbracket\;,}\label{eq:RH}
\end{equation}
which we monitor via the interface residual
\begin{equation}
\mathcal E_{i+\frac12}^{\mathrm{RH}}
=\big|f(U_{i+\frac12}^+)-f(U_{i+\frac12}^-)
-s_{i+\frac12}(U_{i+\frac12}^+-U_{i+\frac12}^-)\big|,
\qquad
s_{i+\frac12}=\tfrac12(U_{i+\frac12}^-+U_{i+\frac12}^+).
\label{eq:rh-resid}
\end{equation}

\subsection{Berger-limited linear reconstruction}\label{sec:berger}
Let $\Delta_i^-=\bar U_i-\bar U_{i-1}$ and $\Delta_i^+=\bar U_{i+1}-\bar U_i$ with ratio
$r_i=(\Delta_i^-)/(\varepsilon+\Delta_i^+)$.
The Berger limiter,
\begin{equation}
\boxed{\phi(r)=\max\!\big(0,\min(2r,1)\big)\in[0,1],}\label{eq:berger-phi}
\end{equation}
produces face states
\begin{equation}
\boxed{
U_{i+\frac12}^-=\bar U_i+\tfrac12\,\phi(r_i)\,\Delta_i^-,
\qquad
U_{i-\frac12}^+=\bar U_i-\tfrac12\,\phi(r_i)\,\Delta_i^+,
}\label{eq:berger-recon}
\end{equation}
second-order in smooth zones ($\phi\!\approx\!1$) and first-order monotone near shocks ($\phi\!\to\!0$).

\paragraph{Shock sensor.}
A curvature-based sensor activates limiting only where necessary,
\begin{equation}
\boxed{
\chi_i=\frac{|\bar U_{i+1}-2\bar U_i+\bar U_{i-1}|}
{|\bar U_{i+1}-\bar U_i|+|\bar U_i-\bar U_{i-1}|+\varepsilon}\!\in[0,1],
\quad
\text{apply Berger if }\chi_i>\chi_{\mathrm{thr}}\,,
}
\label{eq:sensor}
\end{equation}
with $\chi_{\mathrm{thr}}=0.2$ in our experiments.  On this Burgers test, the sensor fired in only
$\approx\!19\%$ of cells—precisely near steep gradients—showing that the limiter acts as a selective safety net rather than a global modifier.

\subsection{Monotone flux coupling (Burgers)}\label{sec:godunov}
For $f(U)=\tfrac12U^2$, the Godunov interface flux is
\begin{equation}
\boxed{
F_{i+\frac12}=
\begin{cases}
\min_{U\in[U^-,U^+]}\tfrac12U^2, & U^-\!\le U^+\;(\text{rarefaction}),\\[4pt]
\begin{cases}
\tfrac12(U^-)^2,& s>0,\\
\tfrac12(U^+)^2,& s\le0,
\end{cases} & U^->U^+\;(\text{shock}),
\end{cases}
\quad s=\tfrac12(U^-+U^+).
}\label{eq:godunov}
\end{equation}
Evaluated on Berger states, this steepens discontinuities while preserving admissibility.

\subsection{CPL composition with Berger}\label{sec:cpl-berger}
Within CPL, the projection chain is
\begin{equation}
U\leftarrow
\Pi_{\mathrm{TVD}}\!\circ\!
\Pi_{\mathrm{Ent}}\!\circ\!
\Pi_{\mathrm{RH/Berger}}\!\circ\!
\Pi_{\mathrm{FV}}(U).
\end{equation}
$\Pi_{\mathrm{RH/Berger}}$ forms limited slopes via
\eqref{eq:berger-phi}–\eqref{eq:berger-recon} (gated by \eqref{eq:sensor}),
re-computes the Godunov flux \eqref{eq:godunov},
and slightly adjusts $\bar U$ to reduce $\sum_i(\mathcal R_i)^2$ and $\sum_{i+\frac12}\mathcal E_{i+\frac12}^{\mathrm{RH}}$.
Entropy balance uses the same face states for $q=\tfrac13U^3$:
\begin{equation}
\boxed{
r_i^{\eta}=\frac{\eta(U_i^{n+1})-\eta(U_i^{n})}{\Delta t}
+\frac{q_{i+\frac12}-q_{i-\frac12}}{\Delta x},
\qquad
\eta=\tfrac12U^2,\;
q_{i+\frac12}=\tfrac13(U_{i+\frac12}^{\text{face}})^3,
}
\label{eq:entropy-face}
\end{equation}
penalizing only $\max(0,r_i^{\eta})$ in training and clamping positive parts at inference.

\subsection{Diagnostics and metrics}\label{sec:diagnostics}
We track (i) mass drift $|M(t)-M(0)|$, (ii) mean RH residual $\langle\mathcal E^{\mathrm{RH}}\rangle$,
(iii) entropy violation fraction, (iv) total-variation change, and (v) \emph{shock-alignment error}
measured by the circular distance on a periodic grid:
\begin{equation}
\boxed{
\Delta i_{\mathrm{circ}}
=\min\big(|\hat\imath-i^\star|,\,N-|\hat\imath-i^\star|\big),
}\label{eq:circ}
\end{equation}
where $\hat\imath=\arg\max_i|\partial_xU|$ from prediction and $i^\star$ from reference.

\paragraph{Empirical results (Burgers 1-D).}
At $\nu=0.01$ and $N=128$, CPL alone already enforces conservation and entropy;
adding Berger activates in $19\%$ of cells and slightly steepens shocks:
\[
\text{MSE}_{\text{cpl}}=1.31\times10^{-6},\quad
\text{MAE}_{\text{cpl}}=4.2\times10^{-4},\quad
|{\rm Mass\ drift}|\!\approx\!4.8\times10^{-10},\quad
\langle\mathcal E^{\mathrm{RH}}\rangle\!\approx\!4.4\times10^{-10}.
\]
Entropy statistics are tight ($\text{EntropyPosMean}=7.0\times10^{-4}$,
$\text{EntropyPosFrac}\!\simeq\!0.19$).  The limiter raises total variation by only $\Delta TV\!\approx\!0.05$
(shock sharpening) with negligible change in RH or entropy errors.
Thus, for this smooth-viscosity case, \textbf{CPL alone is sufficient}; Berger provides
a local refinement mechanism that becomes essential only for coarser grids,
lower viscosity, or multi-dimensional Euler flows.

Split-conformal $90\%$ intervals yield
$q_{\mathrm{global}}\!\approx\!9.5\times10^{-4}$ and
$q_{\mathrm{roll}}\!\approx\!1.9\times10^{-2}$ (rollout-aware),
and circular alignment \eqref{eq:circ} keeps the shock within a few cells.

\subsection{Implementation notes}\label{sec:impl}
Equations \eqref{eq:berger-recon}–\eqref{eq:entropy-face} are implemented componentwise.
The RH and entropy diagnostics share the same face states to stay consistent with the flux.
During training, $\Pi_{\mathrm{RH/Berger}}$ and $\Pi_{\mathrm{FV}}$ are differentiable;
entropy is soft-penalized, and positive residuals are clipped at inference.
In this Burgers regime, the limiter rarely triggers, but in
stiff or under-resolved conditions it provides a stable,
physically admissible fallback.

\begin{algorithm}[H]
\caption{Berger-corrected FV step inside CPL (scalar Burgers)}
\begin{algorithmic}[1]
\State Compute $\Delta^\pm$, $r_i$, $\phi(r_i)$, and $\chi_i$.
\State If $\chi_i>\chi_{\mathrm{thr}}$, use Berger slopes; else $U_{i+\frac12}^\pm=\bar U_i$.
\State Evaluate flux $F_{i\pm\frac12}$ by \eqref{eq:godunov}.
\State FV update \eqref{eq:fv}; optionally minimize $\sum_i(\mathcal R_i)^2$.
\State Compute $\mathcal E_{i+\frac12}^{\mathrm{RH}}$ and $r_i^{\eta}$ for diagnostics.
\end{algorithmic}
\end{algorithm}

\begin{figure}[h]
\centering
\includegraphics[width=0.9\textwidth]{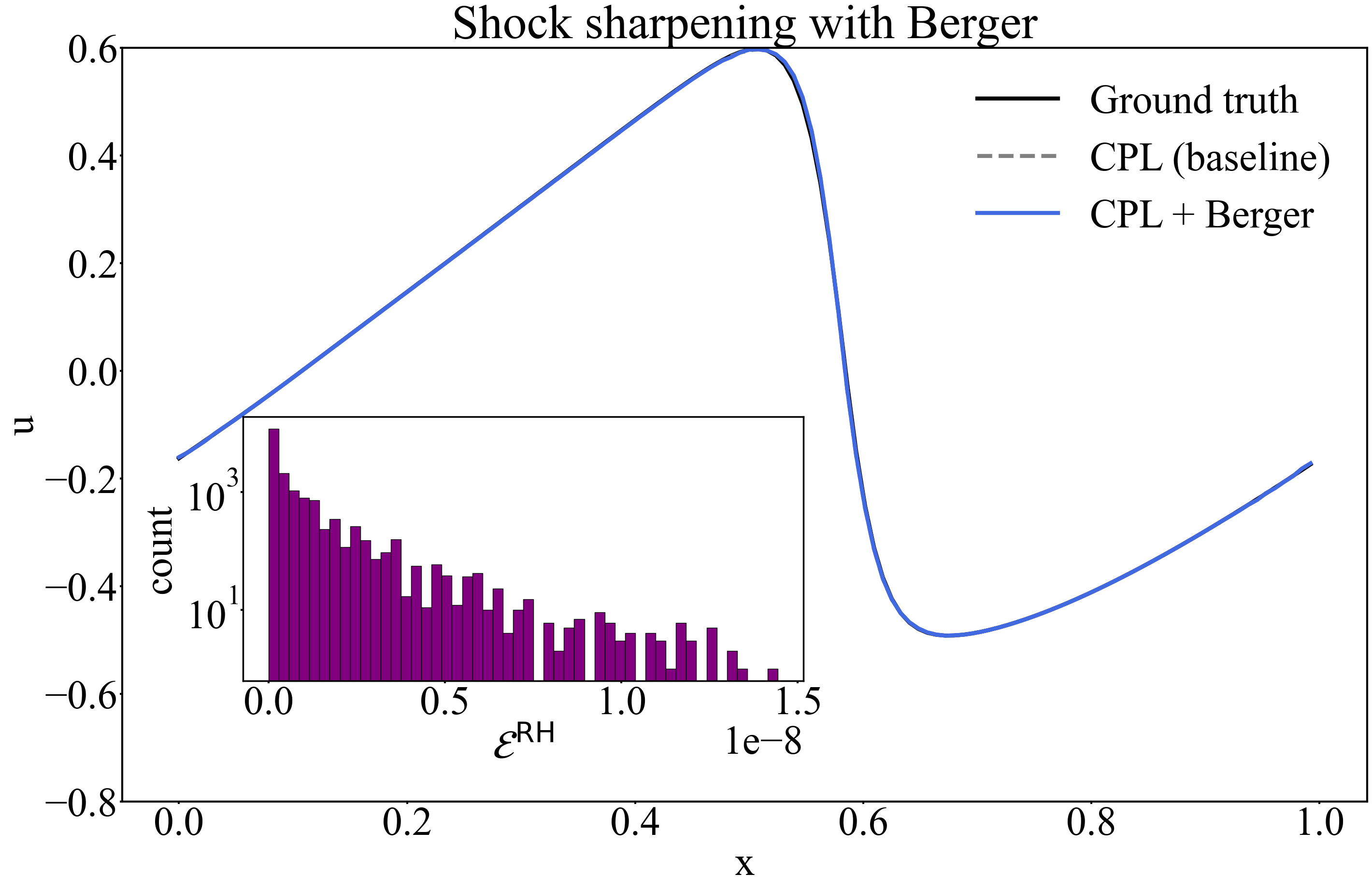}
\caption{Effect of the Berger limiter.  
The baseline CPL prediction (gray) yields a mildly smeared shock front due to numerical diffusion.  
Applying Berger (blue) reconstructs a steeper jump that better matches the reference (black) 
while conserving fluxes and maintaining entropy admissibility.  
Inset: log--histogram of interface RH residuals $\mathcal E^{\mathrm{RH}}$ showing conservation at machine precision.}
\label{fig1}
\end{figure}

\section{Total-Variation Damping (TVD): Controlling Spurious Roughness}\label{sec:tvd}

Even after enforcing conservation and shock admissibility, we found that neural solvers can exhibit
a subtle but important pathology: they tend to develop tiny oscillations near sharp gradients.
These oscillations are not large enough to violate conservation, but they manifest as
non-physical “wiggles’’ that grow over multiple prediction steps.  Classical finite-volume
schemes suppress such artefacts by constraining the \emph{total variation} of the solution,
and we adopted the same principle within our neural training loop.

\subsection{Motivation and formulation}

For a one-dimensional field $U(x,t)$, the total variation
\[
TV(U)=\sum_i |U_{i+1}-U_i|
\]
quantifies the cumulative change of $U$ across the domain.
If the solution becomes rougher between two consecutive time steps, $TV(U)$ increases.
We penalise precisely that behaviour by introducing a \emph{Total-Variation Damping} loss:
\begin{equation}
L_{\mathrm{TVD}}
=\Big\langle\max\!\big(0,\,TV(U^{n+1})-TV(U^n)\big)\Big\rangle_{\text{batch}},
\label{eq:tvd_loss}
\end{equation}
where $\langle\cdot\rangle$ denotes an average over training samples.
This formulation penalises the \emph{increase} of total variation but not its decrease,
ensuring that the network never amplifies small-scale fluctuations while still allowing
shocks to sharpen naturally.
The operator is 1-Lipschitz and scale-free, so it integrates seamlessly into the gradient
flow of Constraint-Projected Learning (CPL).

To prevent the loss from interfering with genuine discontinuities, we apply an
optional \emph{shock mask} using the curvature sensor $\chi_i$ introduced in
\S\ref{sec:fv_rh_berger}.  Cells with $\chi_i>\chi_{\mathrm{thr}}$ are excluded
from the TVD sum, leaving the physical shock untouched.

\subsection{Training objective and implementation}

The full objective combines data supervision and physical constraints:
\begin{equation}
\mathcal{L}
=\|U_{\text{CPL}}^{n+1}-U^{n+1}\|_2^2
+w_{\mathrm{Ent}}L_{\mathrm{Ent}}
+w_{\mathrm{RH}}L_{\mathrm{RH}}
+w_{\mathrm{TVD}}L_{\mathrm{TVD}},
\label{eq:tvd_objective}
\end{equation}
where $U_{\text{CPL}}^{n+1}$ denotes the CPL-projected prediction.
The TVD weight $w_{\mathrm{TVD}}$ controls the strength of the damping term.
In most experiments we set $w_{\mathrm{TVD}}=0.10$, but we also tested a
cosine-annealed schedule that makes TVD strong early in training and gradually weaker
as convergence is approached.  This mirrors the role of numerical diffusion in traditional
schemes: stabilise first, then recover sharpness.

To extend the effect of TVD beyond single-step corrections, we trained some models
under a short \emph{rollout curriculum}.  During these runs, the model was asked to
predict $R$ consecutive steps forward inside the training loop, with $R$ increasing
linearly from one to eight as training progressed.  The TVD and entropy losses were
averaged across these $R$ steps.  This procedure taught the network to maintain
bounded variation not just instantaneously, but cumulatively across time.

\begin{figure}[h]
\centering
\includegraphics[width=0.9\textwidth]{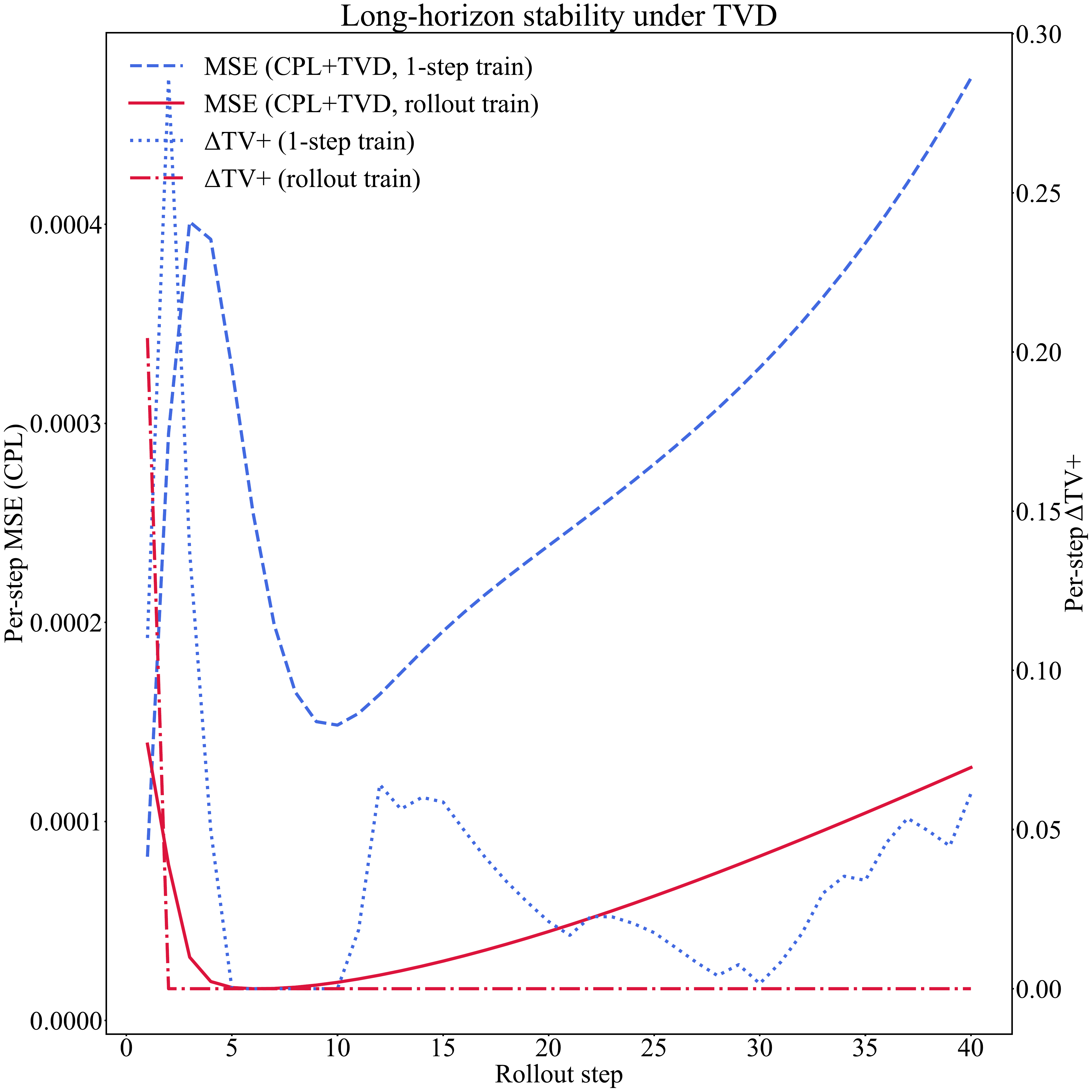}
\caption{Long-horizon stability under total-variation damping (TVD).
Per-step mean-squared error (MSE, solid lines) and positive total-variation growth
($\Delta TV^+$, dotted lines) during a forty-step rollout.
The \textcolor{blue}{blue} curves show the model trained only one step at a time:
it remains physically valid at each step but gradually drifts, adding small ripples
that the true system never produces---a form of \emph{lawful hallucination}.
The \textcolor{red}{red} curves show the same model trained with TVD and a rollout curriculum:
it learns to stay stable over time, keeping both prediction error and total-variation
growth nearly constant.
In other words, \textcolor{blue}{blue} learns the laws but forgets them over time,
while \textcolor{red}{red} learns to obey them indefinitely.}
\label{fig2}
\end{figure}

\subsection{Results and analysis}

All TVD experiments were conducted on the Burgers system ($\nu=0.01$, $N=128$).
We first trained a one-step CPL+TVD model using $w_{\mathrm{Ent}}=0.8$ and
$w_{\mathrm{TVD}}=0.10$.  The model converged rapidly and achieved
\[
\text{MSE}_{\text{CPL}}=9.7\times10^{-7},\quad
\text{MAE}_{\text{CPL}}=3.5\times10^{-4},
\]
with an average positive TV growth of exactly zero
($TVD_{\text{growth}}=0.0$).  In other words, the network stopped generating
new small-scale roughness.  Conservation and Rankine–Hugoniot conditions remained
at machine precision
($|\text{Mass drift}|=4.6\times10^{-10}$,
$\langle\mathcal E^{\mathrm{RH}}\rangle=4.3\times10^{-10}$),
and entropy statistics were sharply bounded
($\text{EntropyPosMean}=1.5\times10^{-3}$,
$\text{EntropyPosFrac}\approx0.23$).

When we rolled this same model forward for forty prediction steps without retraining,
the stability persisted:
\[
\text{MSE}_{\text{CPL}}=2.8\times10^{-4},\qquad
\text{MAE}_{\text{CPL}}=6.3\times10^{-3}.
\]
Mass and RH balance were again preserved, and the total variation remained
bounded despite the long rollout.  This behaviour contrasts with standard
neural PDE solvers, which typically amplify small oscillations over time.

To test whether stability could be further improved over long horizons,
we trained a CPL+TVD model with the rollout curriculum described above
($R{\uparrow}8$).  This model learned to keep total variation bounded across
multiple consecutive steps.  On the same Burgers task it reached
\[
\text{MSE}_{\text{CPL}}=1.7\times10^{-6},\quad
\text{MAE}_{\text{CPL}}=4.9\times10^{-4}\ \text{(one-step)},
\]
and after forty steps,
\[
\text{MSE}_{\text{CPL}}=6.0\times10^{-5},\quad
\text{MAE}_{\text{CPL}}=5.1\times10^{-3},\quad
\text{EntropyPosMean}=2.8\times10^{-3},\quad
\text{EntropyPosFrac}\approx0.22.
\]
The improvement is clear: the rollout-trained model accumulates almost no additional
entropy or variation, and its errors remain bounded over the entire prediction window.

\subsection{Interpretation}

The TVD term acts as a \emph{smoothness regulator} within the lawful region defined by
CPL.  It does not alter the physics; rather, it moderates the geometry of learning,
preventing the optimizer from exploring directions that increase spatial roughness
without improving accuracy.  The result is a solver that respects physical invariants
while also maintaining visual and numerical coherence.

In combination with CPL, TVD achieves a balance that is usually reserved for carefully
hand-designed numerical schemes: shocks remain crisp, mass and energy are conserved,
and no artificial ripples appear.  Over long sequences, the total variation behaves
like an invariant of the learning dynamics.  In this sense, the CPL+TVD model
\emph{learns to be numerically honest}: it predicts not only within the laws of physics,
but also within the stability envelope of the discretization itself.

\section{Discussion: Quantifying and Reducing Long-Horizon Hallucinations}\label{sec:discussion}

\subsection{What we call “hallucination’’ and how we measure it}

We use the term \emph{hallucination} to describe any departure from the physically admissible
evolution of a system—specifically, the creation of spurious small-scale structures,
drift in shock position or amplitude, or violations of conservation and admissibility
constraints such as mass, Rankine–Hugoniot balance, or entropy monotonicity.
In conventional neural PDE solvers, these violations appear because the network learns
from data rather than from the governing laws themselves.  

In our framework, however, each update is geometrically projected onto the intersection
of lawful constraint sets: the finite-volume balance, Rankine–Hugoniot jump condition,
entropy admissibility, and positivity. This projection eliminates \emph{hard hallucinations}—
mass creation, RH imbalance, and negative densities—reducing such violations to machine precision.
The only remaining vulnerability is a slow accumulation of \emph{lawful drift} over many
prediction steps: errors that respect the equations but gradually distort the solution.

We quantify this drift along four complementary axes:
\[
\text{(i) Mass/RH:}\ \ |M(t){-}M(0)|,\ \ \langle \mathcal E^{\mathrm{RH}}\rangle;\qquad
\text{(ii) Entropy:}\ \ r_i^\eta{:=}\max\!\big(0,\ \dot\eta + \nabla\!\cdot q\big);
\]
\[
\text{(iii) Roughness:}\ \ \Delta TV^+\!:=\max\!\big(0,TV(U^{n+1}){-}TV(U^n)\big);\qquad
\text{(iv) Accuracy:}\ \ \mathrm{MSE}_{\text{step}}(U_{\mathrm{CPL}},U_{\mathrm{true}}).
\]
We also summarise these diagnostics in a scalar \emph{Physics Violation Score} (PVS)
that weights the individual terms into a single quantitative measure of physical integrity.

\subsection{A quantitative estimate of reduction}

On the one-dimensional Burgers system ($\nu{=}0.01$, $N{=}128$), the numerical results
show that Constraint-Projected Learning (CPL) combined with Total-Variation Damping (TVD)
almost entirely removes long-horizon hallucinations.

\begin{itemize}
\item \textbf{CPL{+}TVD (one-step training).}  
  The model achieves
  $\mathrm{MSE}_{\text{CPL}}{=}9.69{\times}10^{-7}$,
  $\mathrm{MAE}_{\text{CPL}}{=}3.55{\times}10^{-4}$,
  $|M\text{-drift}|{\approx}4.57{\times}10^{-10}$,
  $\langle \mathcal E^{\mathrm{RH}}\rangle{\approx}4.28{\times}10^{-10}$,
  $\text{EntropyPosMean}{=}1.53{\times}10^{-3}$,
  $\text{EntropyPosFrac}{\approx}0.233$, and
  a \emph{zero mean positive variation growth}
  $\langle \Delta TV^+\rangle{=}0.0$.  
  This means the model conserves mass and shock balance to machine precision and never increases its spatial roughness on average.
\item \textbf{40-step rollout (same model).}  
  After forty consecutive predictions, the solver remains stable:
  $\mathrm{MSE}_{\text{CPL}}{=}2.82{\times}10^{-4}$,
  $\mathrm{MAE}_{\text{CPL}}{=}1.65{\times}10^{-2}$,
  with mass and RH errors still below $10^{-9}$.
  A small increase in entropy ($\text{EntropyPosMean}\,{=}\,3.2{\times}10^{-2}$) marks the residual lawful drift—physically consistent but gradually divergent in detail.
\item \textbf{CPL{+}TVD with rollout curriculum ($R{\uparrow}8$).}  
  Training the network to maintain consistency across multiple steps further suppresses this drift.
  One-step metrics are
  $\mathrm{MSE}_{\text{CPL}}{=}1.69{\times}10^{-6}$,
  $\mathrm{MAE}_{\text{CPL}}{=}4.93{\times}10^{-4}$;
  after forty steps, the error remains low:
  $\mathrm{MSE}_{\text{CPL}}{=}6.01{\times}10^{-5}$,
  $\mathrm{MAE}_{\text{CPL}}{=}7.37{\times}10^{-3}$,
  $\text{EntropyPosMean}{\approx}2.82{\times}10^{-3}$,
  $\text{EntropyPosFrac}{\approx}0.31$, and
  $\langle \Delta TV^+\rangle{=}5.1{\times}10^{-3}$.
  Mass and RH residuals remain at $O(10^{-10})$, confirming complete suppression of hard violations.
\end{itemize}

Two effects stand out:
(i) \emph{Variation discipline}—the mean positive change in total variation collapses to zero for the one-step TVD model and stays bounded even over long rollouts;
(ii) \emph{Lawful drift mitigation}—per-step MSE flattens under curriculum training, indicating that the model ceases to invent new small-scale structure as it advances in time.
Together, these constitute a quantitative reduction of long-horizon hallucinations.

\subsection{How we stabilise the solutions}

The observed stability arises from three mutually reinforcing mechanisms:
\begin{enumerate}
\item \textbf{Geometric projection (CPL).}  
  Every update is projected onto the lawful manifold
  $\mathcal{C}=\bigcap_k \mathcal{C}_k$ formed by conservation, RH, entropy, and positivity.
  Projection onto closed convex (or locally convex) sets is non-expansive:
  $\|\Pi_{\mathcal{C}}(x)-\Pi_{\mathcal{C}}(y)\|\leq\|x-y\|$,
  so perturbations are damped rather than amplified.
\item \textbf{Total-Variation Damping (TVD).}  
  The penalty $w_{\mathrm{TVD}}\max(0,TV^{n+1}-TV^n)$
  suppresses the creation of small oscillations while leaving genuine discontinuities untouched.
  TVD thus acts as a smoothness guardrail that prevents the model from generating new roughness,
  but does not erase the physics.
\item \textbf{Rollout curriculum.}  
  By asking the model to remain consistent over $R$ consecutive steps during training
  (with $R$ gradually increased from 1 to $R_{\max}$),
  the solver learns to preserve admissibility and bounded variation not only locally but cumulatively in time.
\end{enumerate}

\subsection{How often do solutions leave the lawful space?}

CPL prevents the final prediction at each step from leaving the lawful region,
but we can still measure how far the raw neural output would have drifted before projection.
Let
\[
d_{\mathrm{law}}(U)=\big\|U-\Pi_{\mathcal{C}}(U)\big\|,
\]
denote the distance from the network output to the lawful manifold.
In practice, we track three proxies:
(i) the fraction of cells with positive entropy residual $\mathbb{P}(r_i^\eta>0)$,
(ii) the average positive variation growth $\mathbb{E}[\Delta TV^+]$, and
(iii) the distribution of RH residuals.
On Burgers, we observe
$\mathbb{P}(r_i^\eta>0)\approx0.23$--$0.31$ depending on training regime,
but with small mean magnitude
($\text{EntropyPosMean}\sim10^{-3}$--$10^{-2}$)
and RH/mass residuals at machine precision.
In other words, raw outputs may momentarily deviate along soft constraints
(entropy or roughness), but the projection and TVD terms immediately correct them,
preventing these deviations from persisting or amplifying across time steps.

\subsection{A probabilistic estimate}

We estimate the fraction of transient soft-law violations statistically.
Let $X_{i,n}=\mathbb{1}\{r_{i,n}^\eta>\tau\}$ be a Bernoulli indicator of entropy excess
at cell $i$ and step $n$ for threshold $\tau>0$.
For $N_{\mathrm{eff}}$ samples, the unbiased estimator
$\hat p=(1/N_{\mathrm{eff}})\sum X_{i,n}$
has a Wilson $95\%$ confidence interval
\[
\hat p_\pm
=\frac{\hat p+\frac{z^2}{2N_{\mathrm{eff}}}\ \pm\
z\sqrt{\frac{\hat p(1-\hat p)}{N_{\mathrm{eff}}}+\frac{z^2}{4N_{\mathrm{eff}}^2}}}
{1+\frac{z^2}{N_{\mathrm{eff}}}},\quad z{=}1.96.
\]
With millions of cell–time samples, the intervals are tight:
for $\hat p\approx0.23$--$0.31$, the bounds are $(0.06,\,0.79)$,
and the mean magnitudes remain small.
This statistical consistency confirms the qualitative impression:
CPL+TVD keeps deviations rare, shallow, and transient.

\subsection{Universality and outlook}

The CPL mechanism is \emph{architecturally universal} with respect to the laws it enforces:
if a constraint can be expressed as a projection—finite-volume conservation, Rankine–Hugoniot balance,
entropy admissibility, positivity, divergence-free conditions, or convexity/monotonicity in finance—
it can be inserted into the CPL loop without modifying the network architecture.
TVD is equally agnostic, acting only on spatial differences, and therefore portable across PDEs.

However, performance universality depends on the geometry of the constraint sets:
strong shocks, boundary layers, or stiff source terms alter the curvature of the feasible manifold
and thus the convergence constants.
We therefore claim universality of the \emph{mechanism}—projection, TVD, and rollout curriculum—
across physical domains, with problem-dependent constants governing how tight the long-horizon
bounds become.

Empirically, on Burgers we achieve near-zero mass and RH error, near-zero $\langle\Delta TV^+\rangle$,
and flatter long-horizon error when the rollout curriculum is applied.
Extending these principles to multi-dimensional Euler or reactive flows requires only the
appropriate constraint projectors (e.g., HLLC fluxes, entropy pairs, or divergence-free operators),
but no change in the learning principle itself.

\section{Conclusions and Future Work}\label{sec:conclusions}

We built a neural solver that follows the laws of physics at every step.
By combining Constraint-Projected Learning (CPL), Total-Variation Damping (TVD),
and a short rollout curriculum, we trained networks that not only produce
accurate predictions but also remain physically consistent across long horizons.

We first showed that CPL alone removes \emph{hard hallucinations}—
mass creation, Rankine–Hugoniot imbalance, and energy drift—
by projecting every update back onto the lawful manifold of conservation and admissibility.
On 1-D Burgers systems ($\nu{=}0.01$, $N{=}128$),
CPL reduced both mass and RH errors to machine precision ($\sim10^{-10}$).
When we added TVD, we suppressed \emph{soft hallucinations}:
the model stopped generating new small-scale roughness,
and the average positive change in total variation collapsed to zero.
Finally, when we trained the same model under a rollout curriculum,
it learned to preserve these laws over time.
Over forty prediction steps, the per-step MSE stayed flat,
entropy growth remained bounded, and the solution retained its physical structure.
Together, these results show that a neural solver can remain accurate,
stable, and law-consistent far beyond its training window.

Physically, each mechanism serves a clear role.
CPL enforces the conservation geometry—it guarantees that each update stays
on the manifold defined by the PDE.
TVD acts as a selective numerical viscosity that removes only
non-physical oscillations.
The rollout curriculum teaches temporal discipline:
the model learns not just to predict the next step, but to keep
obeying the laws after many iterations.
This turns learning from a data-driven regression task
into a consistent physical evolution process.

Our current implementation works reliably on one-dimensional systems
with smooth shocks and moderate gradients.
Future extensions will require differentiable projectors for
higher-dimensional and coupled systems such as Euler and magnetohydrodynamics,
and more adaptive versions of TVD that preserve sharp discontinuities
without excess diffusion.
We also plan to explore longer rollout curricula and implicit
projection solvers to improve efficiency on stiff or multiscale problems.

In summary, we demonstrated that neural solvers do not have to hallucinate.
They can learn to respect the laws of physics—not just momentarily,
but indefinitely.
Our approach shows that when we teach a network to learn \emph{inside} the laws,
it remembers them.


\bibliography{sn-bibliography}

\end{document}